\documentclass[runningheads]{llncs}
\usepackage{makecell}
\usepackage[utf8]{inputenc}
\usepackage{booktabs}    
\usepackage{multirow}   
\usepackage{enumitem}
\usepackage{graphicx}
\usepackage{amsmath}    
\usepackage{float}
\usepackage[table]{xcolor}
\definecolor{light-gray}{gray}{0.92} 
\usepackage{newunicodechar}
\newunicodechar{，}{,}
\usepackage{stfloats}
\usepackage{xcolor}
\makeatletter
\def\@fnsymbol#1{\ensuremath{%
  \ifcase#1\or *\or \dagger\or \ddagger\or
  \mathsection\or \mathparagraph\or \|\or **\or
  \dagger\dagger\or \ddagger\ddagger\else\@ctrerr\fi}}
\makeatother

\usepackage{eccv}

\usepackage{eccvabbrv}

\usepackage[accsupp]{axessibility}  
\usepackage{hyperref}

\usepackage{orcidlink}

\begin{document}

\title{VisNec: Measuring and Leveraging Visual Necessity for Multimodal Instruction Tuning}
\titlerunning{VisNec: Measuring and Leveraging Visual Necessity}
\authorrunning{M. Dong et al.}

\author{
    Mingkang Dong\inst{1,2}\orcidlink{0009-0000-0276-6075}\thanks{Equal contribution.} \and
    Hongyi Cai\inst{2}\protect\footnotemark[1]\orcidlink{0009-0002-9423-0749} \and
    Jie Li\inst{1}  \orcidlink{0009-0003-2055-7674} \and 
    Sifan Zhou\inst{3}\orcidlink{CMU 0000-0003-3602-7566} \and\\
    Bin Ren\inst{4}\orcidlink{0000-0002-9790-1504} \and
    Kunyu Peng\inst{5}\orcidlink{0000-0002-5419-9292} \and
    Yuqian Fu\inst{1,6}\orcidlink{0000-0002-0412-5500}\thanks{Corresponding author: Yuqian Fu was a visiting scholar at SJTU during this work.}
}

\institute{
    $^1$SJTU,
    $^2$Universiti Malaya,
    $^3$CMU,
    $^4$MBZUAI,
    $^5$KIT, 
    $^6$KAUST
}

\maketitle

\begin{figure*}[h]
\centering
\includegraphics[width=0.97\textwidth]{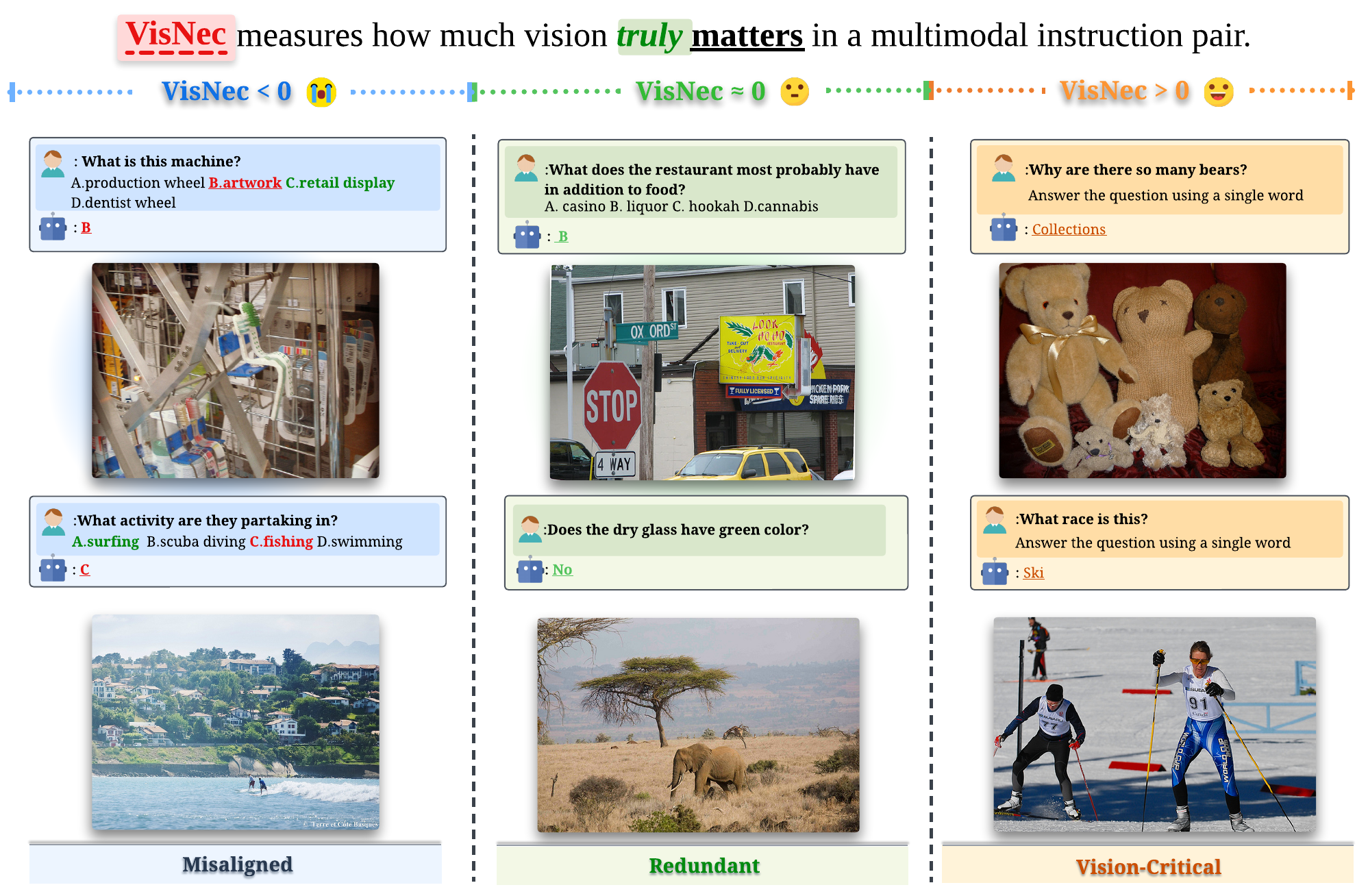}
\caption{
\textbf{Illustration of the VisNec score}, which measures the contribution of visual input in multimodal instruction tuning.
Samples are categorized as \textit{Misaligned} (VisNec < 0, visual input degrades prediction), \textit{Redundant} ($\approx 0$, visual input provides no additional benefit), or \textit{Vision-Critical} (> 0, visual input substantially improves prediction).
}
\label{fig:teaser}
\end{figure*}
\begin{abstract}
The effectiveness of multimodal instruction tuning depends not only on dataset scale, but also critically on whether training samples genuinely require visual reasoning. However, existing instruction datasets often contain a substantial portion of visually redundant samples (solvable from text alone), as well as multimodally misaligned supervision that can degrade learning. To address this, we propose \textbf{VisNec (Visual Necessity Score)}, a principled data selection framework that measures the marginal contribution of visual input during instruction tuning.
By comparing predictive loss with and without visual context, VisNec identifies whether a training instance is vision-critical, redundant, or misaligned. To preserve task diversity, we combine VisNec with semantic clustering and select high-necessity samples within each cluster.
Across 10 downstream benchmarks, training on only 15\% of the LLaVA-665K dataset selected by VisNec achieves 100.2\% of full-data performance. On the smaller Vision-Flan-186K dataset, our selection not only further reduces data size but surpasses full-data training by 15.8\%. These results demonstrate that measuring and leveraging visual necessity provides an effective solution for both efficient and robust multimodal instruction tuning. \textbf{Project Page: \textcolor{RoyalBlue}{https://dmk041218.github.io/VisNec/}}.
\end{abstract}

\section{Introduction}
\label{sec:intro}
Multimodal instruction tuning~\cite{shen2024multimodalinstructiontuningconditional, liu2023visualinstructiontuning, vision-g1, vift} plays a central role in training multimodal large language models (MLLMs), enabling them to follow complex vision-language instructions and generalize across diverse tasks~\cite{singh2019vqa, hudson2019gqa,shu2025earthmind, sigma, yu, liang, OmniEgo-R, fu2025objectrelator, zheng2025multimodal, zhang2025egonight, pan2025v2sammarryingsam2multiprompt, tan2025xtrack, li2025egocross, lu2025yo, wen2025earth, wen2025micamultiagentindustrialcoordination}. Beyond architectural design, the quality of instruction data fundamentally determines whether models develop genuine cross-modal reasoning or rely on superficial correlations. Consequently, curating high-quality multimodal instruction data is critical for robust and reliable performance.

While large-scale instruction datasets~\cite{pistachio} collected from diverse sources have successfully trained strong MLLMs, directly using all available data presents two notable limitations. 
First, it is inefficient. As datasets scale to hundreds of thousands or millions of samples, training costs grow substantially, yet not all samples contribute equally to learning.
Second, large-scale instruction data inevitably contains a significant proportion of redundant or misaligned samples. 
Visually redundant instances can often be answered through linguistic shortcuts alone (e.g., predicting ``green'' for ``the color of grass''), providing little incentive for visual grounding. When such samples dominate training, models tend to exploit textual correlations rather than integrate visual evidence, weakening cross-modal dependence. 
In contrast, multimodal misalignment, arising from annotation errors or noisy data pipelines, produces image-text pairs with inconsistent supervision. Training on such samples can actively degrade visual reasoning and amplify hallucinations at inference time.
These observations raise a fundamental question:
\textit{Can we identify a small yet truly valuable subset of multimodal instruction data that is both efficient and effective?}

Existing data selection approaches primarily rely on generic importance or diversity signals, such as gradient influence~\cite{wu2025iconsinfluenceconsensusvisionlanguage}, transferability~\cite{naharas2025dataselectionfinetuningvision}, or clustering-based coverage~\cite{lee2024coincide,ofa}. However, these methods treat multimodal samples holistically and do not explicitly distinguish the independent contribution of the visual modality. As a result, selected subsets may still favor linguistically easy samples or retain harmful misaligned supervision. In contrast, we posit that the value of a multimodal instruction sample largely depends on whether visual input is necessary for resolving predictive uncertainty. This perspective shifts the focus from generic sample importance to \textbf{visual necessity} as the guiding principle for multimodal data selection.

To implement this principle, we draw inspiration from the theory of V-usable information~\cite{ethayarajh2025understandingdatasetdifficultymathcalvusable}, which formalizes how much a variable reduces predictive uncertainty beyond what is already available from other contexts. While V-usable information has been used to assess textual data quality, extending this principle to multimodal instruction data remains underexplored.  We argue that this perspective provides the right lens for evaluating visual supervision: a training instance is only as valuable as the uncertainty that can be uniquely resolved by visual input. 
Grounded in this insight, we introduce \textbf{Visual Necessity Score (VisNec)}, a lightweight data selection framework that explicitly quantifies the marginal contribution of the visual modality. By contrasting model losses with and without visual input, VisNec produces a principled per-sample measure of visual necessity. As shown in Fig.~\ref{fig:teaser}, VisNec successfully identifies instances that are vision-critical, redundant, or misaligned.  Moreover, considering that visual dependency varies across task types, e.g., geometric reasoning naturally demands stronger visual grounding than knowledge-based QA, we integrate stratified semantic clustering with intra-cluster VisNec ranking. This ensures that the selected subset maintains both visual indispensability and broad task coverage.

Extensive experiments across ten benchmarks demonstrate that VisNec recovers or even surpasses full-data performance using only 15\% of the training data, achieving 100.2\% on LLaVA-665K~\cite{liu2023visualinstructiontuning} and 115.8\% on Vision-Flan-186K~\cite{xu-etal-2024-vision}. Besides, our method generalizes across model scales from 3B to 32B.
Our main contributions are summarized as follows:

\begin{enumerate}
    \item We identify a critical yet overlooked limitation in existing multimodal data selection: the neglect of the visual modality's independent contribution, which leads to widespread ``pseudo-multimodal'' samples that reinforce linguistic shortcuts and hinder true cross-modal reasoning.
    \item We propose {VisNec}, a lightweight and model-relative data selection framework that quantifies the marginal contribution of visual input, ensuring the selected subset is both visually indispensable and task-diverse.
    \item We achieve state-of-the-art performance across diverse training datasets and benchmarks, demonstrating that measuring visual necessity enables both data efficiency and improved multimodal reasoning robustness.
\end{enumerate}

\section{Related Work}
\textbf{Multimodal Instruction Tuning.} To enable Multimodal Large Language Models (MLLMs) to effectively handle diverse downstream tasks, instruction tuning has emerged as a fundamental stage in the training pipeline. Established frameworks such as LLaVA~\cite{liu2023visualinstructiontuning}, InstructBLIP~\cite{dai2023instructblipgeneralpurposevisionlanguagemodels}, and Qwen-VL~\cite{bai2023qwenvlversatilevisionlanguagemodel} demonstrate that fine-tuning on a broad range of instruction-following pairs is critical for aligning multimodal representations with human intent. However, the effectiveness of this alignment typically relies on the availability of large-scale, \mbox{high-quality} datasets. As these collections scale to hundreds of thousands of samples, the computational resources and training time required for fine-tuning become significant bottlenecks. Furthermore, 
large-scale instruction datasets, whether automatically collected or model-generated, inevitably include redundant and misaligned samples, which can weaken visual supervision and potentially degrade multimodal reasoning.

\textbf{Information-Theoretic Data Quality.} 
A fundamental question in data-centric learning is how to quantify the value of individual training samples. V-usable information~\cite{ethayarajh2025understandingdatasetdifficultymathcalvusable} formalizes this as the reduction in predictive uncertainty contributed by a variable beyond existing context, providing a principled view of feature-level informativeness. Building on this perspective, dataset difficulty measures~\cite{EL2N} and loss-based scoring methods~\cite{li2024quantityqualityboostingllm} treat predictive loss as an indicator of sample quality in LLM fine-tuning. These works demonstrate that loss-based signals are effective in single-modal settings. However, by operating on text alone, existing approaches cannot disentangle visual informativeness from linguistic ambiguity. A sample may appear difficult due to annotation noise or textual complexity rather than genuine visual contribution. By contrast, our work extends this information-theoretic perspective to multimodal instruction tuning by explicitly isolating the marginal contribution of visual input through counterfactual loss comparison, thus providing a modality-aware quality signal.

\textbf{Data Selection for Instruction Tuning.} Existing data selection methods suffer from a range of practical limitations. Complexity-driven approaches like Self-Filter~\cite{chen-etal-2024-vision} rely on multi-scoring mechanisms that are themselves computationally expensive. Methods such as PreSel~\cite{safaei2025filterimagesfirstgenerate} and CoIDO~\cite{yan2025coidoefficientdataselection} offload quality judgment to closed-source APIs, incurring both financial costs and privacy concerns. COINCIDE~\cite{lee2024coincide} and XMAS~\cite{naharas2025dataselectionfinetuningvision} improve scalability through clustering and SVD-based alignment trajectories, respectively, yet still impose notable overhead on large-scale datasets. Gradient-based methods like ICONS~\cite{wu2025iconsinfluenceconsensusvisionlanguage} offer precise importance estimation at the cost of per-sample backpropagation, making them prohibitively slow in practice. Others are directly adapted from the NLP paradigm~\cite{cai2025mergeitselectionmergingefficient,cai2025lowconfidencegoldrefininglowconfidence}, applying language-centric quality metrics that inherently favor samples with rich textual content but weak visual grounding. IFD~\cite{li2024quantityqualityboostingllm}, most closely related to our work, selects samples by their response prediction loss under text-only input. However, this conflates genuine visual complexity with textual ambiguity or annotation noise. A misaligned sample may score high on IFD precisely because the image misleads the model, yet such samples are detrimental to training. VisNec resolves this by computing the differential between blind and multimodal losses, explicitly rewarding visual informativeness while penalizing misalignment. 
 Overall, existing data selection methods typically suffer from limited computational efficiency, dependence on external resources, or insufficient sensitivity to cross-modal quality, 
 making truly efficient multimodal instruction tuning difficult in resource-constrained settings.
 
\section{Methodology}
\textbf{Problem Formulation.} Let $\mathcal{D} = \{(v_i, t_i, y_i)\}_{i=1}^N$ 
denote a multimodal instruction tuning dataset, where $v_i$ represents the input 
image, $t_i$ the textual instruction, and $y_i$ the target response. The objective 
for fine-tuning a Multimodal Large Language Model (MLLM), parameterized by $\theta$, 
is to minimize the multimodal training loss $\mathcal{L}_{\text{MM}}$, defined as 
the negative log-likelihood of the response $y$ given the multimodal context:
\begin{equation}
    \mathcal{L}_{\text{MM}}(\theta; v, t, y) = -\sum_{j=1}^{L} \log P(y_j \mid y_{<j}, v, t; \theta),
\end{equation}
where $L$ is the length of the response tokens and $y_{<j}$ denotes the tokens 
preceding position $j$. While effective, blindly optimizing this objective over 
massive datasets often leads to inefficiency, as the model may overfit to linguistic 
priors in samples where visual information is redundant. A fundamental question 
arises: which samples genuinely require visual reasoning, and which are solvable 
by linguistic priors alone? Equation~(1) does not differentiate between these 
cases, treating all samples equally, regardless of their visual necessity.
\subsection{Visual Necessity Score (VisNec)}
\textbf{Towards Visual-Centric Data Selection.}
Grounded in the V-usable information framework~\cite{ethayarajh2025understandingdatasetdifficultymathcalvusable}, 
which quantifies how much a variable $X$ reduces predictive uncertainty about $Y$ via:
\begin{equation}
    I_{\mathcal{V}}(X \rightarrow Y) = H_{\mathcal{V}}(Y) - H_{\mathcal{V}}(Y \mid X),
\end{equation}
where $H_{\mathcal{V}}(Y)$ and $H_{\mathcal{V}}(Y \mid X)$ denote the V-entropy 
without and with input $X$ respectively, we argue that the value of a multimodal 
training sample should be measured by how much the visual modality uniquely reduces 
predictive uncertainty beyond what text alone provides. This motivates a direct 
approximation within an MLLM: we quantify visual necessity as the marginal loss 
reduction attributable to the visual input:
\begin{equation}
    \Delta\mathcal{L}(v, t, y) = \mathcal{L}(y \mid t) - \mathcal{L}(y \mid t, v).
\end{equation}
A larger $\Delta\mathcal{L}$ indicates higher visual necessity, meaning the visual input substantially reduces predictive uncertainty beyond what text alone provides. Conversely, a smaller or negative $\Delta\mathcal{L}$ suggests 
the sample is either linguistically solvable or multimodally misaligned. We address the concrete computation of $\mathcal{L}(y \mid t)$ within an MLLM in the following section.\\
\textbf{The Blind Forward Pass.}
To instantiate $\mathcal{L}(y \mid t)$ within a multimodal model, we perform a 
counterfactual inference called the \textit{Blind Forward Pass}. For each sample 
$(v, t, y)$, we replace image token positions with padding tokens and suppress 
their attention contributions by setting the corresponding attention mask to zero, 
ensuring that no visual information can propagate into response token 
representations through any attention computation. This keeps the model 
architecture and parameter space identical to the standard forward pass, with 
the presence of visual input as the sole controlled variable between the two passes. 
To ensure that both passes compute loss over identical response token positions, 
image token positions are excluded from loss computation via \texttt{ignore\_index}, 
such that $\mathcal{L}_{\text{Blind}}$ and $\mathcal{L}_{\text{MM}}$ are both 
averaged exclusively over response tokens:
\begin{equation}
    \mathcal{L}_{\text{Blind}}(\theta; t, y) = -\sum_{j=1}^{L} \log P(y_j \mid y_{<j}, t; \theta).
\end{equation}
$\mathcal{L}_{\text{Blind}}$ thus quantifies the model's residual uncertainty in 
predicting $y$ from linguistic context $t$ alone. Although the blind input lies 
outside the MLLM's training distribution, this does not undermine the validity 
of VisNec as a selection signal: since data selection operates via intra-cluster 
ranking rather than absolute score thresholds, any systematic bias introduced 
by the out-of-distribution input cancels out in the relative ordering.\\
\textbf{Visual Necessity Score.}
We define the VisNec score $S_{\text{VisNec}}$ as the discrepancy between the 
blind loss and the multimodal loss, representing the marginal contribution of 
the visual modality to the model's prediction:
\begin{equation}
    S_{\text{VisNec}}(v, t, y) = \mathcal{L}_{\text{Blind}}(\theta; t, y) - \mathcal{L}_{\text{MM}}(\theta; v, t, y).
\end{equation}
This formulation offers a clear interpretation of sample utility:
\begin{itemize}[label=$\bullet$, leftmargin=0.4cm, nosep]
    \item $S_{\text{VisNec}} > 0$: \textbf{High Visual Gain.} The image significantly 
    reduces prediction error, indicating the sample requires true cross-modal reasoning.
    \item $S_{\text{VisNec}} \approx 0$: \textbf{Redundancy.} The model predicts 
    the answer equally well without the image, suggesting the sample relies heavily 
    on linguistic priors.
    \item $S_{\text{VisNec}} < 0$: \textbf{Misalignment/Noise.} The presence of 
    the image increases the loss, implying the visual content contradicts the text 
    or introduces noise.
\end{itemize}

\subsection{Semantic-Aware Stratified Sampling}
Relying solely on VisNec scores for selection carries a risk: the distribution of 
visual dependency is non-uniform across tasks. For instance, geometric reasoning 
tasks naturally have higher VisNec scores than OCR tasks. A naive top-$k$ selection 
might bias the dataset toward specific domains. To preserve broad task diversity, 
we adopt a coarse-to-fine paradigm that first groups samples by semantic intent 
and then performs intra-cluster selection.\\
\textbf{Instruction Clustering.}
To capture the true semantic structure of the dataset, we focus on the core task 
definition encapsulated in the user's query. Specifically, we extract the question 
component $q_i$ from each textual instruction $t_i$ (filtering out system prompts) 
and map it into a semantic embedding space. We then perform K-Means clustering to 
learn a partition function $\pi: q \rightarrow \{1, \dots, K\}$, which assigns 
each question to one of $K$ semantic clusters. This explicitly groups samples 
based on \textit{task intent}, isolating distinct capabilities such as geometric 
reasoning, OCR, and creative generation.\\
\textbf{Intra-Cluster Selection.}
We construct the $k$-th semantic cluster $\mathcal{C}_k = \{ (v_i, t_i, y_i) \in 
\mathcal{D} \mid \pi(q_i) = k \}$. Within each cluster, we first remove samples 
with $S_{\text{VisNec}} \leq 0$, as these indicate either misaligned image-text 
pairs where the visual input actively misleads the model or redundant samples 
where the visual context contributes nearly nothing to the response. The remaining 
samples are then ranked by their VisNec scores. Given a budget ratio $r$, we 
select the top-$r\%$ instances to form the final subset:
\begin{equation}
    \mathcal{D}_{\text{select}} = \bigcup_{k=1}^{K} \text{Top-}{r\%} 
    \left( \{ (v_i, t_i, y_i) \in \mathcal{D} \mid \pi(q_i) = k \} ; S_{\text{VisNec}} \right).
\end{equation}

\begin{figure*}[!h]
  \centering
  \includegraphics[width=\linewidth]{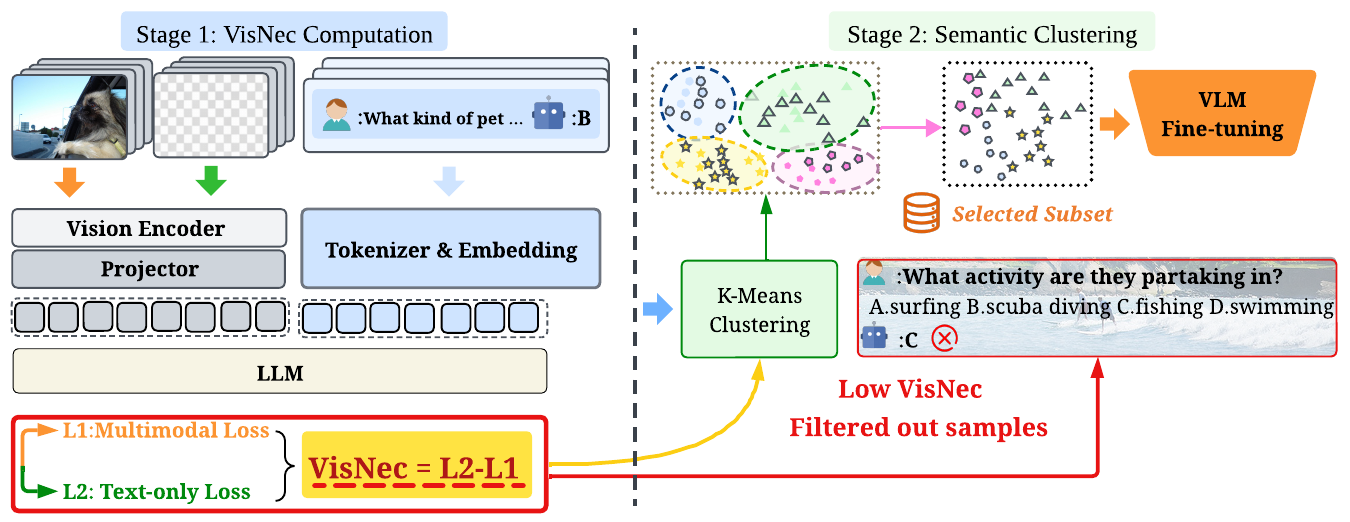}
  \caption{\textbf{Overview of the VisNec Data Selection Framework.} Each sample 
  undergoes two forward passes to compute its VisNec score. Samples with non-positive 
  scores are filtered out, and the remainder are grouped via K-Means clustering, 
  from which the top-$r$\% are selected within each cluster for fine-tuning.}
  \label{fig:pipeline}
\end{figure*}

\noindent\textbf{Overall Framework.} As illustrated in Fig.~\ref{fig:pipeline}, our framework 
operates in two stages. In the first stage, each sample is scored by computing 
the difference between its text-only loss and multimodal loss within the target 
MLLM, yielding a per-sample Visual Necessity Score that measures how much the visual input actually contributes to the response. Samples with non-positive scores 
are discarded as they are redundant or misaligned, while the rest are ranked by their VisNec scores. In the second 
stage, we perform semantic clustering over the instructions and apply intra-cluster 
selection based on these scores, ensuring that the final subset is both visually indispensable and task-diverse.\\

\subsection{Case Study}
\begin{figure*}[h]
  \centering
  \includegraphics[width=\linewidth]{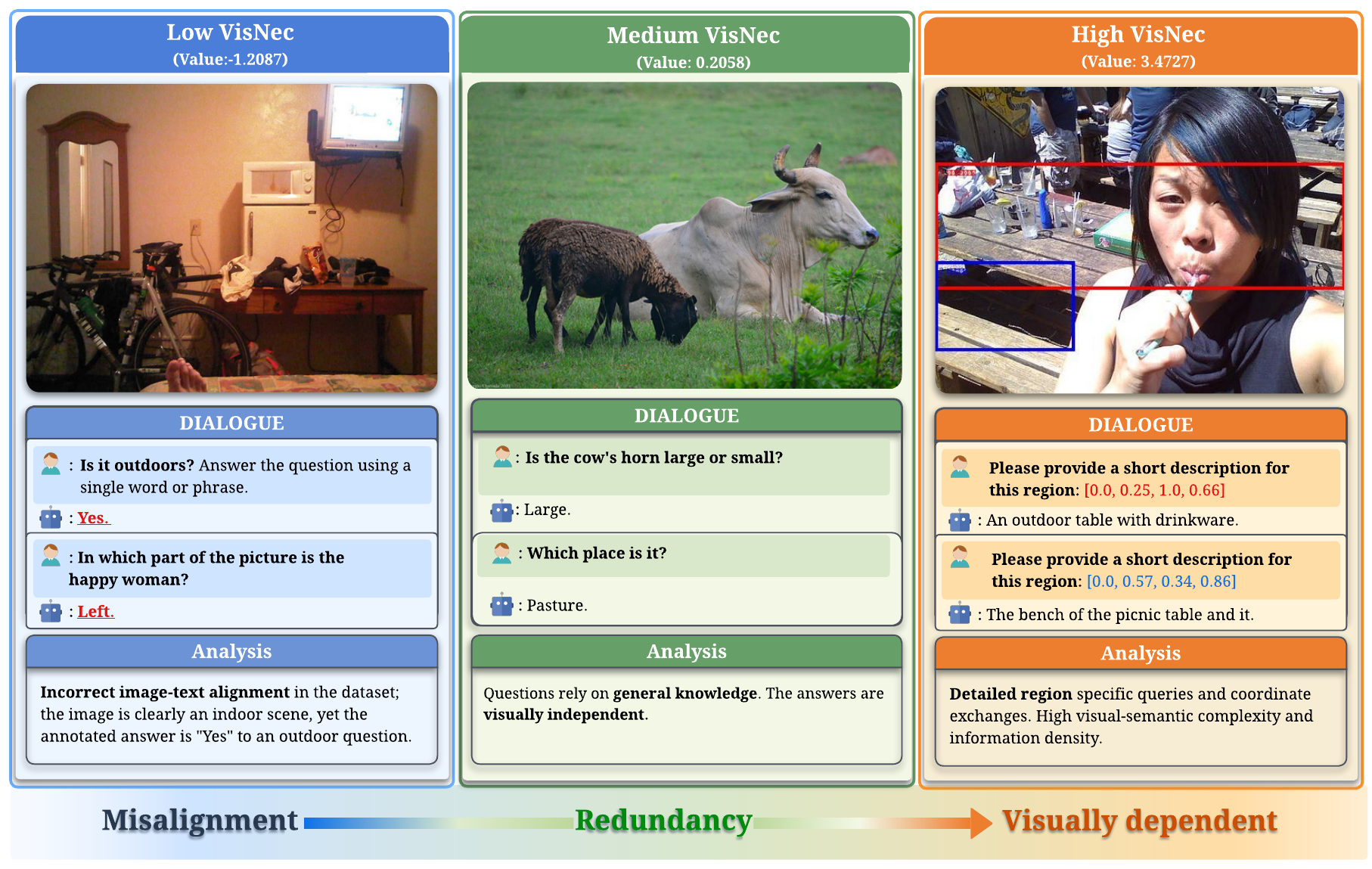}
  \caption{Qualitative comparison of Low ($-1.2087$), Medium($0.2058$) and High VisNec ($3.4727$) samples, illustrating misaligned, redundant and visual dependent samples.}
  \label{fig:case_study}
\end{figure*}

To illustrate how VisNec identified problematic and high-value samples, we show three representative cases in Fig.~\ref{fig:case_study}.
\\
\textbf{Low VisNec (Misalignment).}
The left example receives a negative score (-1.21), revealing multiple annotation errors. First, the question "Is it outdoors?" is answered "Yes," but the image clearly shows an indoor scene with furniture and walls. Second, when asked "In which part of the picture is the happy woman?" the answer is "Left"—yet no woman is visible in the image at all. These contradictions mean the model performs better by ignoring the image entirely. Without visual input, it can at least avoid being misled by annotations that do not match the actual content. VisNec correctly assigns a negative score to such samples and discards them outright, as their visual content actively contradicts the textual supervision rather than complementing it.\\
\textbf{Medium VisNec (Redundant).} The middle example scores 0.21, indicating that the visual input contributes little beyond what linguistic priors already provide. Questions such as "Is the cow's horn large or small?" and "Which place is it?" can largely be answered through general world knowledge alone, without any visual grounding. Such samples are not harmful but offer minimal cross-modal supervision, as the model learns nothing it could not have inferred from text alone. VisNec correctly assigns them a near-zero score and deprioritizes them during selection, effectively treating them as visually redundant.\\
\textbf{High VisNec (Valuable).} The right example scores 3.47, showing strong visual dependency. The 
questions request descriptions of specific regions defined by coordinates 
[0.0, 0.25, 1.0, 0.66] and [0.0, 0.57, 0.34, 0.86]. Without the image, 
these questions are impossible to answer since no amount of linguistic prior knowledge can determine what objects occupy those spatial locations. Such samples require leveraging visual grounding to identify the objects at these specific coordinates.
By filtering out negative scores and prioritizing high scores within each 
semantic cluster, VisNec removes noisy annotations and text-solvable 
shortcuts while retaining samples that teach cross-modal reasoning.
\section{Experiments and Results}
\subsection{Experimental Setup}
\textbf{Datasets.} To validate the effectiveness and scalability of our proposed approach across various magnitudes of instruction-tuning data, we conduct experiments on two datasets: LLaVA-665K and Vision-Flan-186K. LLaVA-665K is a widely used dataset, comprising more than 600K samples
across about 10 different tasks. Vision-Flan-186K includes 191 distinct tasks, which allows us to assess the method's performance when dealing with a high number of tasks and a relatively smaller quantity of data per task. For both experiments, we employ a 15\% sampling ratio  to evaluate performance under budget-constrained scenarios.\\
\textbf{Baselines.} We evaluate our approach against a diverse set of competitive baselines, ranging from traditional sampling to recent state-of-the-art (SOTA) methods, including Random, Self-Filter, EL2N~\cite{EL2N},
TypiClust~\cite{hacohen2022activelearningbudgetopposite},
IFD~\cite{li2024quantityqualityboostingllm},
PreSel~\cite{safaei2025filterimagesfirstgenerate},
XMAS~\cite{naharas2025dataselectionfinetuningvision},
COINCIDE~\cite{lee2024coincide},
CLIP-Score~\cite{radford2021learningtransferablevisualmodels},
and CoIDO~\cite{yan2025coidoefficientdataselection}.
Specifically, XMAS, PreSel, and CoIDO represent recent SOTA techniques in multimodal data selection. In contrast, IFD and EL2N are representative methods originally developed for instruction tuning or data selection in the NLP / supervised-learning setting.\\
\textbf{Evaluation Benchmarks.} We evaluate the performance of different fine-tuned MLLMs across 10 multimodal evaluation benchmarks that test various capabilities of MLLMs. The benchmarks include: 1) Visual Question Answering: VQAV2\cite{goyal2017making}, GQA\cite{hudson2019gqa};   2) Knowledge-Grounded QA: SQA-I\cite{lu2022learn}; 3) Multiple-choice understanding:
MM-Bench (EN/CN)\cite{liu2023mmbench} and MME-Perception\cite{fu2025mmecomprehensiveevaluationbenchmark}; 4) Text understanding in images: TextVQA\cite{singh2019vqa}; 5) Open-ended generation: LLaVA-Wild Bench\cite{liu2023visualinstructiontuning}; 6) Hallucination and factual tests: POPE\cite{li-etal-2023-evaluating}, MM-Vet\cite{yu2024mmvetevaluatinglargemultimodal}.\\
\textbf{Training Details.}
To ensure a fair and rigorous comparison, we strictly adhere to the official training configurations across all experiments, including model architectures, hyperparameters, and optimization schedules. Specifically, starting from the LLaVA-v1.5\cite{liu2023visualinstructiontuning} Stage-1 pre-trained checkpoint (feature alignment), we conduct LoRA fine-tuning for 1 epoch with a default learning rate of $2\times10^{-4}$, following the standard LLaVA training pipeline. For the K-means clustering involved in our approach, we set the number of clusters to $K=20$. The main experimental results are summarized in Tab.~\ref{tab:llava_7b} and Tab.~\ref{tab:vision-flan}, which report the performance comparison between our approach and several state-of-the-art baselines on the LLaVA-665K and Vision-Flan-186K datasets, respectively. Given the varying scales across different benchmarks, we report the Average Relative Performance (Rel.) to facilitate a unified comparison. For each benchmark, relative performance is defined as: 
\begin{equation}
    \text{Rel} = \frac{\text{The Model's Performance}}{\text{Full Fine-tuned Model's Performance}} \times 100\%.
\end{equation}
This metric represents the percentage of the full-data baseline performance achieved by the model trained on the selected data subset.
\subsection{Main Results on Multi-modal Data Selection}
\noindent\textbf{\noindent\textbf{Superiority Over State-of-the-Art Baselines.}}
Tab.~\ref{tab:llava_7b} presents a quantitative comparison of VisNec against competitive baselines for fine-tuning LLaVA-v1.5-7B. Under a strict 15\% data budget on the LLaVA-665K dataset, VisNec achieves a remarkable relative performance of 100.2\%, consistently surpassing all baseline approaches. Specifically, VisNec outperforms the Random baseline by 5.8\% and the second-best method by 2.3\%. We attribute this to a fundamental distinction in how VisNec defines sample value: rather than relying on generic importance signals or external quality proxies, VisNec explicitly computes the difference between blind and multimodal losses, directly measuring how much the visual input reduces predictive uncertainty. This allows VisNec to prioritize vision-critical samples while discarding those where the image actively misleads the model—a dimension that existing methods do not explicitly capture. Notably, VisNec even exceeds the full-data (100\%) fine-tuning baseline on LLaVA-Wild, MMBench-EN/CN, and MM-Vet, benchmarks that demand open-ended reasoning and hallucination resistance. These benchmarks are precisely where noisy and visually redundant training samples are most damaging, and where removing them yields the greatest benefit.\\
\begin{table*}[t]
\centering
\caption{Overall performance and efficiency comparison of selection approaches for fine-tuning LLaVA-v1.5-7b on the LLaVA-665K dataset using  a 15\% (98K samples) sampling ratio across various multimodal evaluation benchmarks. "Rel" indicates relative performance. The best result is \textbf{bolded} and the runner-up is \underline{underlined}.}
\label{tab:llava_7b}
\renewcommand{\arraystretch}{1.0}
\scalebox{0.65}{
\begin{tabular}{clcccccccccccc}
\toprule
& \textbf{Method} & \textbf{VQAv2} & \textbf{GQA} & \textbf{\makecell{\textbf{LLaVA} \\ \textbf{Wild}}} & \textbf{SQA-I} & \textbf{TextVQA} & \textbf{MME-P} & \makecell{\textbf{MMBench} \\ \textbf{en}} & \makecell{\textbf{MMBench} \\ \textbf{cn}} & \textbf{POPE} & \textbf{MM-Vet} & \textbf{Rel.} \\
\midrule
\multicolumn{13}{c}{\textbf{LLaVA-v1.5-7B on LLaVA-665K}} \\
\midrule
\rowcolor{blue!12}
0 & Full-Data(665K) & 79.1 & 63.0 & 67.9 & 68.4 & 57.9 & 1476.9 & 64.3 & 58.3 & 86.4 & 30.0 & 100\% \\
\midrule
1 &  Random& 75.3 & 55.1 & 58.8 & 67.8 &  54.3 & 1397.5 & 61.0 & 53.5 & 84.9 & 30.2& 94.2\%\\
2 &  Self-Filter \scalebox{0.7}{[ACL-2024]} &  74.0 & 56.3& 60.6 & 62.3 & 51.4 & 1356.5 &  48.1 & 45.4 & \underline{86.3} & 29.0 & 89.3\% \\
3 & EL2N \scalebox{0.7}{[NeurIPS-2021]} & 76.1 & 54.7 & 59.0 & 66.5 & 50.2 & 1405.2 & 58.2 & 48.5 & 83.3 & 30.0 & 91.9\% \\
4 &  TypiClust\scalebox{0.7}{[ICML-2022]} & 76.0 & 59.8 & 65.2& 68.2 & 53.3 & 1396.2 & 64.3 & \underline{57.1} & 85.6 & 29.7 & 96.9\% \\
5 &  IFD \scalebox{0.7}{[NAACL-2024]}& 74.0 &57.8 & 62.3 &  66.5 & 51.8 & 1307.2 & 57.2 & 50.6 & \textbf{86.6} & 28.1 & 92.1\% \\
6 &  PreSel \scalebox{0.7}{[CVPR-2025]} &  \underline{76.5} &  57.9 & 65.6 & \textbf{70.1}& 55.2 &  \underline{1457.7} & \underline{64.8} & 56.5 & 85.4 & 29.6 & \underline{97.7\%} \\
7 &  XMAS \scalebox{0.7}{[arxiv-2025]}& 75.1 & \textbf{61.2} & 62.4& 67.0 & 55.2 & \textbf{1485.7} & 64.3 & 54.1 & 85.8 & 31.0 & 97.3\%\\
8 &  COINCIDE\scalebox{0.7}{[EMNLP-2024]} & 76.1 & 60.2 & 64.9 & 67.7 &  54.8 & 1414.9 & 60.5 &53.9 & 86.4 &28.5 & 95.8\% \\
9 &  CLIP-Score\scalebox{0.7}{[ICML-2021]} &   71.7 & 56.9 & 61.3&  64.5 &  53.4 & 1380.3 & 51.0 & 48.0 & 84.0& 29.9 & 92.0\% \\
10 & CoIDO\scalebox{0.7}{[NeurIPS-2025]}  &75.8  &61.0 & \underline{66.7} & \underline{68.2} & \underline{56.0} & 1419.5 & 63.3 & 55.5 & 85.6 & \underline{31.4} & \underline{97.7}\% \\
11 & ICONS\scalebox{0.7}{[arxiv-2025]}  & 76.3  &60.7 & 65.3 & 65.3 & 55.2 & 1435.6 & 63.1 & 55.8 & 85.7& 30.4 & 97.1\%\\
\midrule
\rowcolor{orange!18}12 & VisNec(ours) & \textbf{78.0}& \underline{60.8} &\textbf{69.8} & 67.9 &\textbf{56.2} & 1457.2 & \textbf{64.9} & \textbf{59.1} & 86.0 & \textbf{32.1} & \textbf{100.2\%} \\
\bottomrule
\end{tabular}%
}
\end{table*}

\noindent\textbf{Generalization Across Datasets.} Given that the LLaVA-1.5 
dataset covers a relatively limited range of tasks, we evaluate the cross-dataset generalization of our approach on the more challenging Vision-Flan-186K dataset, which spans 191 distinct tasks. As reported in Tab.~\ref{tab:vision-flan}, VisNec achieves an exceptional relative performance of 115.8\%, outperforming the Random baseline by 18.4\% and the full-data fine-tuned model by 15.8\%. This result reveals a critical insight: a substantial portion of Vision-Flan-186K is not merely redundant but \textbf{harmful} to model training. Samples with low or negative VisNec scores, where visual content contradicts textual annotations, actively degrade cross-modal reasoning. By explicitly filtering such misaligned samples, VisNec confirms that in multimodal instruction tuning, data quality is far more critical than data quantity.
\begin{table*}[t]
\centering
\caption{Overall performance and efficiency comparison of selection approaches for fine-tuning LLaVA-v1.5-7b on the Vision-Flan-186K dataset using  a 15\% (28K samples) sampling ratio across various multimodal evaluation benchmarks. "Rel" indicates relative performance. The best result is \textbf{bolded} and the runner-up is \underline{underlined}.}
\label{tab:vision-flan}
\renewcommand{\arraystretch}{1.0}
\scalebox{0.65}{
\begin{tabular}{clcccccccccccc}
\toprule
& \textbf{Method} & \textbf{VQAv2} & \textbf{GQA} &\makecell{\textbf{LLaVA} \\ \textbf{Wild}} & \textbf{SQA-I} & \textbf{TextVQA} & \textbf{MME-P} & \makecell{\textbf{MMBench} \\ \textbf{en}} & \makecell{\textbf{MMBench} \\ \textbf{cn}} & \textbf{POPE} & \textbf{MM-Vet} & \textbf{Rel.} \\
\midrule
\multicolumn{13}{c}{\textbf{LLaVA-v1.5-7B on Vision-Flan-186K}} \\
\midrule
\rowcolor{blue!12}
0 & Full-Data(186K) & 69.6 & 46.0 & 35.7 & 55.6 & 38.3 & 1238.1 & 53.4 & 48.2 & 85.7 & 27.7 & 100\% \\
\midrule
1 &  Random& \underline{66.5} &  43.8& 33.5& 62.1 &  38.7 & 1238.6 & 43.6 & 43.1 & 83.0 & 28.1& 96.7\% \\
2 &  Self-Filter \scalebox{0.7}{[ACL-2024]} &  64.9 & 42.5& 32.1 & 59.3 & 42.6 & 1262.2 &  42.1 & 43.8 & 80.9 & 25.1 & 95.0\% \\
3 & EL2N \scalebox{0.7}{[NeurIPS-2021]} & 63.6 & 42.2 & 32.8 &  62.7 & 42.4 & 1253.8 & 44.7 & 37.7 & 79.5 & 27.1 & 95.2\% \\
4 &  TypiClust\scalebox{0.7}{[ICML-2022]} &  65.8 & 43.1 & 30.4& 59.2 & 37.7 &  1194.1 & 32.4 & 45.1 & 81.6&  28.2 & 92.0\% \\
5 &  IFD \scalebox{0.7}{[NAACL-2024]}& 65.0 &42.4 & 29.8 & 57.8 & 42.0 & 1210.9 & 30.4 & 40.8 &  82.6 & 26.9 & 91.6\% \\
6 &  PreSel \scalebox{0.7}{[CVPR-2025]} &   64.1  & 41.9  & \underline{39.4} & \underline{66.2}& 39.7 &  1218.8 & 50.4 & 45.4 & 84.1 & \underline{29.1} & 100.6\% \\
7 &  XMAS \scalebox{0.7}{[arxiv-2025.10]}& 65.5 & \underline{49.4} & 34.2
& 58.1 & 42.4 & 1151.2 & 51.1 & 44.0 & 76.2 & 24.3 & 96.9\% \\
8 &  COINCIDE\scalebox{0.7}{[EMNLP-2024]} & 66.0 &44.5 & 34.6& 63.9 &  33.0& 1184.4 & 49.6 &48.2 & \underline{84.3} &26.1 & 97.1\% \\
9 &  CLIP-Score\scalebox{0.7}{[ICML-2021]} &   63.0 & 41.6 & 31.2&  62.3 &  39.7 & 1058.0 & 37.5 & 44.2 & 82.0&  28.0 & 92.2\% \\
10 & CoIDO\scalebox{0.7}{[NeurIPS-2025]}  &66.7  &46.8&37.6 & \underline{66.2}  & \underline{43.2} & \underline{1298.8} & \underline{51.4}  & \underline{47.3} & \textbf{85.6} & 28.3  & \underline{103.6\%} \\
11 & ICONS\scalebox{0.7}{[arxiv-2025]} &67.2&48.8&35.4&60.2&49.9&1252.5&51.3&45.4&83.0&28.6& 103.2\% \\
\midrule
\rowcolor{orange!18}12 & VisNec(ours) & 65.0& \textbf{57.1} &\textbf{39.5} & \textbf{74.5} & \textbf{51.6} & \textbf{1505.5} & \textbf{62.6} & \textbf{53.1} & 82.0 & \textbf{32.1} & \textbf{115.8\%} \\
\bottomrule
\end{tabular}%
}
\end{table*}
\subsection{Transferability across Architectures and Scales}
To verify whether VisNec captures marginal data value 
rather than model-specific biases, we evaluate 
cross-architecture transferability by applying VisNec 
scoring directly on Qwen2.5-VL~\cite{bai2025qwen25vltechnicalreport} 
family across three scales (3B, 7B and 32B) on the 
LLaVA-665K dataset. Despite significant architectural 
differences from LLaVA-v1.5, the subsets selected by 
Qwen2.5-VL-3B, 7B, and 32B all achieve remarkable performance. As shown in Tab.~\ref{tab:dual_dataset_comparison}, VisNec using only 15\% of the full data achieves 103.8\%, 104.0\%, and 102.4\% of their respective full data performance. This demonstrates that VisNec captures the intrinsic visual necessity of the data rather than model-specific preferences, making it a robust and architecture-agnostic selection strategy for scaling next-generation MLLMs with minimal data overhead.
\begin{table*}[!h]
\centering
\caption{Performance comparison of Qwen2.5-VL (3B, 7B and 32B) trained LLaVA-665k subset versus full data. \textbf{Rel.} indicates relative performance compared to the Full method.}
\label{tab:dual_dataset_comparison}
\renewcommand{\arraystretch}{0.8}
\scalebox{0.67}{
    \begin{tabular}{lccccccccccccc}
    \toprule
    \textbf{Model} & \textbf{Data} & \textbf{Method} & \textbf{VQAv2} & \textbf{GQA} & \makecell{\textbf{LLaVA}\\\textbf{Wild}} & \textbf{SQA-I} & \textbf{TextVQA} & \textbf{MME-P} & \makecell{\textbf{MMB}\\\textbf{(en)}} & \makecell{\textbf{MMB}\\\textbf{(cn)}} & \textbf{POPE} & \textbf{MM-Vet} & \textbf{Rel.} \\
    \midrule
    \multirow{3}{*}{Qwen2.5VL-3B}
    & 665k & Full   & 79.7 & 59.4 & 73.5 & 70.3 & 73.6 & 1475.6 & 72.0 & 69.0 & 87.6 & 50.2 & 100\% \\
    \cmidrule(lr){2-14}
    & \multirow{2}{*}{98k}
    & Random & 78.3 & 58.4 & 72.1 & 71.3 & 72.1 & 1439.2 & 71.9 & 68.3 & 87.0 & 49.1 & 98.8\% \\
    & & VisNec & \textbf{80.7} & \textbf{59.5} & \textbf{75.4} & \textbf{76.0} & \textbf{76.8} & \textbf{1552.5} & \textbf{76.2} & \textbf{75.4} & \textbf{88.3} & \textbf{50.5} & \textbf{103.8\%} \\
    \midrule
    \multirow{3}{*}{Qwen2.5VL-7B}
    & 665K & Full   & 83.4 & \textbf{62.9} & 77.8 & 75.6 & 77.8 & 1592.5 & 81.3 & 78.1 & 87.5 & 50.7 & 100\% \\
    \cmidrule(lr){2-14}
    & \multirow{2}{*}{98k}
    & Random & 82.1 & 61.3 & 77.5 & 75.2 & 79.2 & 1549.3 & 80.2 & 76.2 & 88.1 & 48.5 & 98.7\% \\
    & & VisNec & \textbf{83.3} & 62.1 &\textbf{80.9} & \textbf{87.7} & \textbf{83.7} & \textbf{1603.6} & \textbf{82.8} & \textbf{80.2} & \textbf{88.5} & \textbf{54.3} & \textbf{104.0\%} \\
    \midrule
    \multirow{3}{*}{Qwen2.5VL-32B}
    & 665K & Full   & \textbf{82.1} & 60.2 & 75.5 & 81.9 & 77.8 & 1684.5 & 80.8 & 78.6 & \textbf{88.5} & 52.3 & 100\% \\
    \cmidrule(lr){2-14}
    & \multirow{2}{*}{98k}
    & Random & 80.3 & 58.1 & 74.2 & 80.6 & 75.1 & 1648.2 & 79.5 & 78.2 & 86.3 & 51.6 & 97.9\% \\
    & & VisNec & 81.5 &\textbf{60.6} & \textbf{78.0} & \textbf{86.0} & \textbf{78.2} & \textbf{1687.2} & \textbf{82.2} & \textbf{81.4} & 88.1 & \textbf{57.7} & \textbf{102.4\%} \\
    \bottomrule
    \end{tabular}
}
\end{table*}
\subsection{Parameter Sensitivity Study}
\begin{figure}[h]
    \centering
    \includegraphics[width=\linewidth]{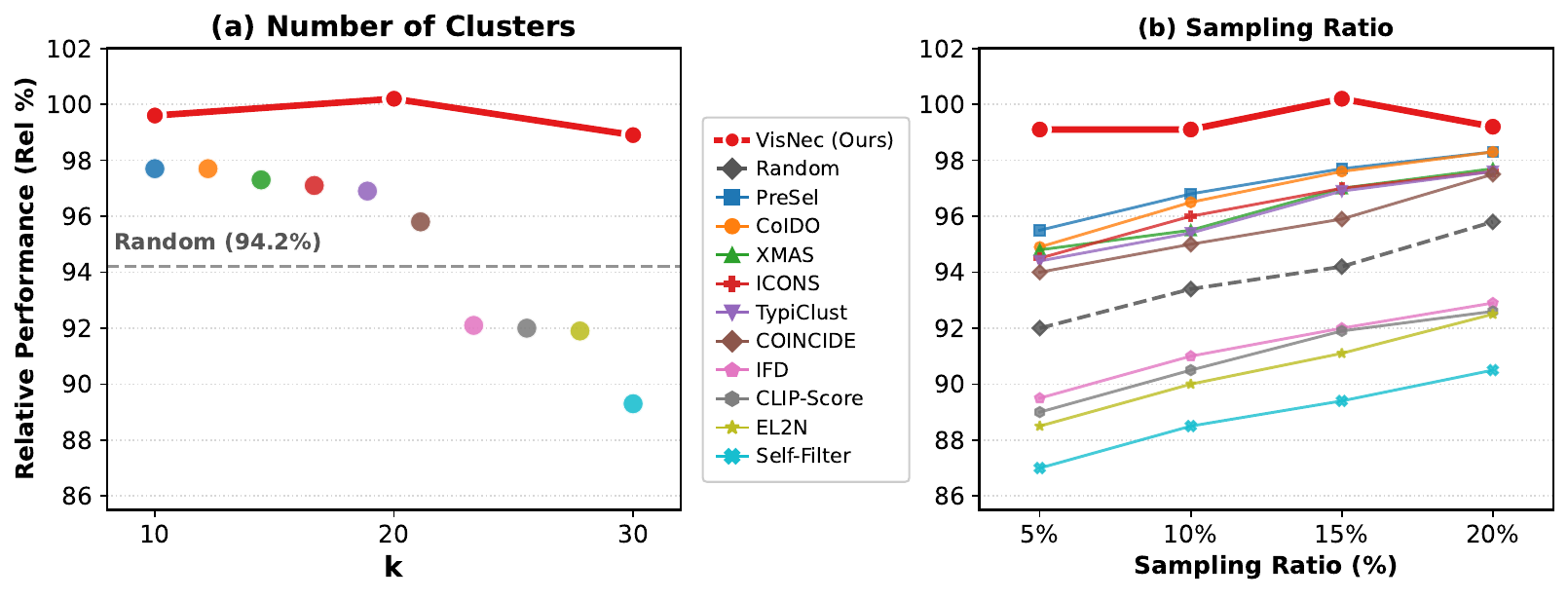}
    \caption{Impact of (a) the number of clusters $k$ in K-means; (b) the sampling ratio $\rho$.}
    \label{fig:sensitivity}
\end{figure}

To evaluate the sensitivity of our proposed method, we conduct an ablation study on two key components: the number of clusters $k$ in K-means and the selection ratio $\rho$.\\
\textbf{Analysis of the Number of Clusters ($k$).} We investigate the impact of the number of clusters $k$ in our stratified sampling strategy. As shown in Fig.~\ref{fig:sensitivity}, our method remains highly stable across a range of $k$ from 10 to 30. The relative performance (Rel) fluctuates within a narrow margin of 1.3\%, consistently maintaining over 98.9\% of the full-data performance and outperforming other methods. This suggests VisNec is robust to different levels of semantic granularity. We use $k=20$ as the default setting in our experiments.\\
\textbf{Analysis of Selection Ratio.} We investigate the impact of the selection ratio ($\rho$) on model performance, sweeping from 5\% to 20\%. As illustrated in Fig.~\ref{fig:sensitivity}, VisNec exhibits a remarkable ability for strategic data concentration. Specifically, using only 5\% of the training data, our method restores 99.1\% of the full-data baseline performance. The performance reaches its peak at $\rho = 15\%$, achieving 100.2\% relative performance, which effectively outperforms the model trained on the entire LLaVA-665K dataset. This result indicates that VisNec goes beyond mere data reduction; it performs a semantic-aware refinement of the instruction set. By leveraging VisNec scores to identify samples with the highest multi-modal information gain within each semantic cluster, VisNec isolates a high-fidelity core-set that maximizes learning efficiency. The slight performance dip at 20\% (99.2\% Rel.) reinforces the "less-is-more" paradigm in visual instruction tuning, confirming that excessive data may introduce gradient noise or redundancy. Consequently, we utilize 15\% as our optimal selection ratio to maintain peak accuracy with minimal computational cost.

\subsection{Contribution of Loss Components}
The core premise of VisNec is that sample value is best captured by the marginal contribution of the visual modality over linguistic context. To validate this, we disaggregate our scoring mechanism and evaluate the individual contributions of each loss component against a Random sampling baseline, as detailed in Tab.~\ref{tab:loss}. Our analysis reveals a clear hierarchy in selection efficacy. The Text-only strategy (95.6\% Rel.) outperforms Random sampling (94.2\%), showing that linguistic difficulty is a reasonable proxy for sample quality. However, the Multimodal-only variant surprisingly falls below the random baseline at 94.0\%. Without the contrastive signal from the text-only loss, the selection process tends to favor samples with high visual complexity but low instructional clarity, effectively introducing noise into the training subset.

In contrast, the full VisNec framework achieves 100.2\% relative performance, a 4.6\% margin over the best single-modality baseline. This confirms that VisNec captures a synergy between textual and visual signals that neither modality alone can provide—identifying samples where the two are mutually informative rather than redundant or conflicting. Consequently, training on just 15\% of VisNec-selected data surpasses full-data supervision, indicating that the differential loss formulation is key to isolating genuinely cross-modal samples.
\begin{table}[!h]
    \centering
    \caption{Ablation study on loss components. We report results on LLaVA-v1.5-7B as the base model. Our default setting is k=20, $\rho = 15\%$.}
    \label{tab:loss}
    \renewcommand{\arraystretch}{0.8}
    \scalebox{0.70}{
        \begin{tabular}{lccccccccccc}
            \toprule
            \textbf{Model} & \textbf{VQAv2} & \textbf{GQA} & \textbf{LLaVA} & \textbf{SQA-I} & \textbf{TextVQA} & \textbf{MME-P} & \textbf{MMBench} & \textbf{MMBench} & \textbf{POPE} & \textbf{MM-Vet} & \textbf{Rel.} \\
             & & & \textbf{Wild} & & & & \textbf{(en)} & \textbf{(cn)} & & & \\
            \midrule
            LLaVA-7B 100\% & 79.1 & 63.0 & 67.9 & 68.4 & 57.9 & 1476.9 & 64.3 & 58.3 & 86.4 & 30.0 & 100\% \\
            \midrule
            Random     & 75.3 & 55.1 & 58.8 & 67.8 & 54.3 & 1397.5 & 61.0 & 53.5 & 84.9 & 30.2 & 94.2\% \\
            Text       & 75.0 & 58.3 & 66.0 & 64.7 & 54.7 & 1436.2 & 61.4 & 55.2 & 86.6 & 28.4 & 95.6\% \\
            Multimodal & 76.2 & 57.5 & 59.7 & 65.4 & 54.1 & 1443.5 & 60.2 & 51.6 & 85.3 & 29.2 & 94.0\% \\
            VisNec     & \textbf{78.0} & \textbf{60.8} & \textbf{69.8} & \textbf{67.9} & \textbf{56.2} & \textbf{1457.2} & \textbf{64.9} & \textbf{59.1} &\textbf{86.0} & \textbf{32.1} & \textbf{100.2\%} \\
            \bottomrule
        \end{tabular}
    }
\end{table}
\subsection{Contribution of Clustering}
We further investigate whether semantic clustering is truly necessary or whether selecting top-r\% samples by VisNec score alone (TopVisNec) suffices. As shown in Tab.~\ref{tab:contribution_of_clustering}, TopVisNec outperforms Random sampling yet still lags behind the full framework by 3.2\%. This gap suggests that VisNec scores alone, without cluster-level constraints, are insufficient to maintain a balanced task distribution across the selected subset.
\begin{table}[!h]
    \centering
    \caption{Ablation study on clustering. We report results on LLaVA-v1.5-7B as the base model. Our default setting is k=20, $\rho = 15\%$.}
    \label{tab:contribution_of_clustering}
    \renewcommand{\arraystretch}{0.8}
    \scalebox{0.70}{
        \begin{tabular}{lccccccccccc}
            \toprule
            \textbf{Model} & \textbf{VQAv2} & \textbf{GQA} & \textbf{LLaVA} & \textbf{SQA-I} & \textbf{TextVQA} & \textbf{MME-P} & \textbf{MMBench} & \textbf{MMBench} & \textbf{POPE} & \textbf{MM-Vet} & \textbf{Rel.} \\
             & & & \textbf{Wild} & & & & \textbf{(en)} & \textbf{(cn)} & & & \\
            \midrule
            LLaVA-7B 100\% & 79.1 & 63.0 & 67.9 & 68.4 & 57.9 & 1476.9 & 64.3 & 58.3 & 86.4 & 30.0 & 100\% \\
            \midrule
            Random     & 75.3 & 55.1 & 58.8 & 67.8 & 54.3 & 1397.5 & 61.0 & 53.5 & 84.9 & 30.2 & 94.2\% \\
            Top-VisNec   & 75.9 & 58.5 & 66.8 & 66.2 & 55.6 & 1383.3 & 63.9 & 56.9 & \textbf{86.2} & 30.0 & 97.0\% \\
            VisNec     & \textbf{78.0} & \textbf{60.8} & \textbf{69.8} & \textbf{67.9} & \textbf{56.2} & \textbf{1457.2} & \textbf{64.9} & \textbf{59.1} & 86.0 & \textbf{32.1} & \textbf{100.2\%} \\
            \bottomrule
        \end{tabular}
    }
\end{table}
\subsection{Cost Analysis}
We evaluate the computational efficiency of VisNec by comparing it with several state-of-the-art data selection baselines on the LLaVA-665K dataset using LLaVA-v1.5-7B. As shown in Tab.~\ref{tab:cost_comparison}, while full-data fine-tuning requires 76.0 GPU-hours, our method VisNec achieves superior performance (100.2\% Rel.) with a total cost of only 23.0 GPU-hours. The data selection process of VisNec takes only 12.0 GPU-hours, which is significantly faster than standard filtering methods like Self-Filter (73.5 GPU-hr) and COINCIDE (55.5 GPU-hr). Unlike PreSel or CoIDO, VisNec does not rely on any external LLM APIs~\cite{openai2024gpt4technicalreport}.
\begin{table}[!h]
\caption{Comparison of computational cost and performance on LLaVA-665K. Time metrics are reported in H100 GPU-hours. ``API Cost'' refers to extra expenditures for external LLM services (e.g., GPT-4). Rel. indicates overall relative performance of the full-data fine-tuned model.}
\label{tab:cost_comparison}
\centering
\renewcommand{\arraystretch}{1.0}
\scalebox{0.85}{
\begin{tabular}{lccccc}
\toprule
Methods & Selection Time & Finetuning Time & API Cost & Total Cost & Rel. \\
\midrule
\rowcolor{blue!12}Full Finetune & --          & 76.0 GPU-hr & \texttimes & 76.0 GPU-hr & 100\% \\
PreSel        & 9.0 GPU-hr  & 11.0 GPU-hr & \checkmark & 20.0 GPU-hr + API Cost & 97.7\% \\
CoIDO         & 25.0 GPU-hr & 11.0 GPU-hr & \checkmark & 36.0 GPU-hr + API Cost & 97.7\% \\
Self-Filter   & 73.5 GPU-hr & 11.0 GPU-hr & \texttimes & 84.5 GPU-hr & 89.3\% \\
XMAS          & 36.0 GPU-hr & 11.0 GPU-hr & \texttimes & 47.0 GPU-hr & 97.3\% \\
COINCIDE      & 55.5 GPU-hr & 11.0 GPU-hr & \texttimes & 66.5 GPU-hr & 95.8\% \\
\rowcolor{orange!18}\textbf{VisNec (Ours)} & 12.0 GPU-hr & 11.0 GPU-hr & \texttimes & 23.0 GPU-hr & \textbf{100.2\%} \\
\bottomrule
\end{tabular}
}
\end{table}
\\
\\
\section{Conclusions}
In this paper, we introduce VisNec (Visual Necessity Score), a data-centric
framework designed to address the redundancy and misalignment prevalent in
current multimodal instruction tuning datasets. By explicitly quantifying the
marginal contribution of visual context relative to linguistic priors, VisNec effectively filters out misaligned and redundant samples while preserving task diversity. Extensive experiments demonstrate that VisNec consistently identifies
high-value, visually indispensable samples that outperform full-data baselines at
a fraction of the computational cost, with strong robustness to hyperparameters
and cross-architecture transferability.

\section{Acknowledgment}
This work is supported by the computing resources provided by TeleAI.
\bibliographystyle{splncs04}
\bibliography{main}
\end{document}